\documentclass[11pt,a4paper]{article}
\usepackage{emnlp2020}
\usepackage{times}
\usepackage{latexsym}

\usepackage{microtype}
\usepackage{tabularx}
\usepackage{amsfonts}
\usepackage{pgfplots}
\usepackage{graphicx}
\usepackage{amsmath}
\usepackage{booktabs}

\aclfinalcopy 

\title{An exploration of the encoding of grammatical gender in word embeddings}

\author{
    Hartger Veeman\\
    Dep. of Linguistics and Philology \\
    Uppsala University\\
    hartger.veeman.7544@student.uu.se
  \And
    Ali Basirat\\
    Dep. of Linguistics and Philology \\
    Uppsala University\\
    ali.basirat@lingfil.uu.se
    
}

\date{}

\begin{document}
\maketitle

\begin{abstract}
The vector representation of words, known as word embeddings, has opened a new research approach in linguistic studies. These representations can capture different types of information about words. The grammatical gender of nouns is a typical classification of nouns based on their formal and semantic properties. The study of grammatical gender based on word embeddings can give insight into discussions on how grammatical genders are determined. In this study, we compare different sets of word embeddings according to the accuracy of a neural classifier determining the grammatical gender of nouns.
It is found that there is an overlap in how grammatical gender is encoded in Swedish, Danish, and Dutch embeddings. Our experimental results on the contextualized embeddings pointed out that adding more contextual information to embeddings is detrimental to the classifier's performance. We also observed that removing morpho-syntactic features such as articles from the training corpora of embeddings decreases the classification performance dramatically, indicating a large portion of the information is encoded in the relationship between nouns and articles.  

\end{abstract}


\section{Introduction}
The creation of embedded word representations has been a key breakthrough in natural language processing \cite{NIPS2013_5021,pennington2014glove,fasttext,elmo,devlin-etal-2019-bert}. Words embeddings have proven to capture information relevant to many tasks within the field.
\citet{linguisticinformation} have shown that word embeddings contain information about the grammatical gender of nouns. In their study, a classifier's ability to classify embeddings on grammatical gender is interpreted as an indication of the presence of information about gender in word embeddings. 
The way this information is encoded in word embeddings is of interest because it can contribute to the discussion of how nouns of a language are classified into different gender categories. 

This paper explores the presence of information about grammatical gender in word embeddings from three perspectives:
\begin{enumerate}
    \item Examine how such information is encoded across languages using a model transfer between languages.
    \item Determine the classifier's reliance on semantic information by using contextualized word embeddings.
    \item Test the effect of gender agreements between nouns and other words categories on the information encoded into word embeddings.
\end{enumerate}

\section{Grammatical gender}
Grammatical gender is a nominal classification system found in many languages. In languages with grammatical gender, the noun forms an agreement with another language aspect based on the noun class. The most common grammatical gender divisions are masculine/feminine, masculine/feminine/neuter, and uter/neuter, but many other divisions exist.

Gender is assigned based on a noun's meaning and form \citep{AliEtAl_inpress}. Gender assignment systems vary between languages and can be based solely on meaning \cite{corbett}.
The experiments in this paper concern grammatical gender in Swedish, Danish, and Dutch. 

Nouns in Swedish and Danish can have one of two grammatical genders, uter, and neuter indicated through the article for indefinite form and the suffix for the definite and plural form. Dutch nouns can technically be classified as masculine, feminine, or neuter. However, in practice, the masculine and feminine genders are only used for nouns referring to people, with the distinction between masculine and feminine being the subject and object pronouns, while in other cases, all non-neuter nouns can be categorized as uter. Grammatical gender in Dutch is indicated in the definite article, adjective agreement, pronouns, and demonstratives. 

\section{Multilingual embeddings and model transferability}
The first experiment explores if grammatical gender is encoded in a similar way between different languages. For this experiment, we leverage the multilingual word embeddings. Multilingual word embeddings are word embeddings that are mapped to the same latent space so that the vector of words that have a similar meaning has a high similarity regardless of the source language.
The aligned word embeddings allow for the model transfer of the neural classifier. This is done by applying the neural classifier model to a different language than it is trained on. If the model effectively classifies embeddings from a different language, it is likely grammatical gender is encoded in the embeddings of two languages similarly.


\subsection{Data \& Experiment}
The lemma form of all unique nouns and their genders were extracted from Universal Dependencies treebanks \cite{nivre16lrec}. This resulted in about 4.5k, 5.5k, and 7.2k nouns in Danish (da), Swedish (sv), and Dutch (nl), respectively. Ten percent of each data set was sampled randomly to be used as test data. For each noun, the embeddings was extracted from pre-trained aligned word embeddings published in the MUSE project \cite{muse}. The uter/neuter class distributions are 74/26  for Swedish, 68/32 for Danish, and 75/25 for Dutch.

The classification is performed using a feed-forward neural network. The network has an input size of $300$ and a hidden layer size of $600$. The loss function used is binary cross-entropy loss. The training is run until no further improvement is observed. 

For every language, a network was trained using the data described above. Every model was then applied to its own test data and the other languages' test data to test the model's transferability.


\begin{table}[h]
\renewcommand{\arraystretch}{1.3}
\caption{Accuracy for model (vertical) applied to test set (horizontal) }
\label{table_example}
\centering
\begin{tabularx}{\columnwidth}{@{}XXXXXX@{}}
\toprule
&   & SV    & DA    & NL   &     \\ \midrule
& SV & 93.55 & 73.89 & 73.37 &  \\
& DA & 81.18 & 91.81 & 78.50& \\
& NL & 71.32 & 78.54 & 93.34 & \\  \bottomrule
\end{tabularx}
\end{table}
\begin{table}[h]
\renewcommand{\arraystretch}{1.3}
\caption{Corrected accuracy for model (vertical) applied to test set (horizontal) }
\label{table_example}
\centering
\begin{tabularx}{\columnwidth}{@{}XXXXXX@{}}
\toprule
&   & SV    & DA    & NL      &  \\ \midrule
&SV & 32.03 & 15.25 & 11.37&  \\
&DA & 22.54 & 35.33 & 19.50& \\
&NL & 9.32 & 19.54 & 31.15 & \\  \bottomrule
\end{tabularx}
\end{table}
\subsection{Results}

To isolate how much of the accuracy of a model is caused by successfully applying learned patterns from the source language to the target language rather than successful random guessing, the following baseline is defined: 
\begin{equation*}
    \sum_{g \in \{u, n\}} p(g_s)p(g_t)
\end{equation*}
This baseline is the chance the model would make a correct guess purely based on the class distributions of the training data and test data. Subtracting this chance from the results should give an indication of how much information in the model was transferable to the task of classifying test data in another language. The absolute results are displayed in table 1 while the results corrected with this baseline are displayed in table 2.  

All transfers yielded a result above the random guessing baseline. This means the classifier is learning patterns in the source language data that are applicable to the target language data. From this, it can be concluded that grammatical gender is encoded similarly between these languages to a certain degree.

The performance of the Swedish model for classifying Dutch grammatical gender and vice versa is the lowest of all language pairs. This language pair is also the pair with the largest phylogenetic language distance. The language pair with the smallest distance is Danish/Swedish. The Danish model applied to the Swedish test data produces the best result of all model transfers, achieving an accuracy of $21.06\%$ over the baseline. This observation on the inverse relationship between the language distance and the gender transferability agrees with the broader experiments performed by \citep{hartger_conll}. 

Interestingly, the success of the model transfer is not symmetrical. This is most clear in the Swedish-Danish pair which achieve an accuracy of $22.54\%$ over baseline from Danish to Swedish, but only $15.25\%$ over baseline from Swedish to Danish. The patterns the Danish model learns are more applicable to the Swedish data than vice versa.

\section{Contextual word embeddings}
Adding contextual information to a word embedding model has proven to be effective for semantic tasks like named entity recognition or semantic role labeling \cite{elmo_layers}. This addition of semantic information can be leveraged to measure its influence on the gender classification task. A contextualized word embedding model was used to quantify the role of semantic information in the noun classification. 
\subsection{Embeddings from Language Models (ELMo)}
The ELMo model \cite{elmo} is a multi-layer biRNN that leverages pre-trained deep bidirectional language models to create contextual word embeddings. The ELMo model has three layers, where layer 0 is the non-contextualized token representation, which is a concatenation of the word embeddings and character-based embeddings created with a CNN or RNN. This token representation is fed into a biRNN. Afterward, the resulting hidden state is concatenated with the states of both directions of the language model and fed into another biRNN layer.

\cite{elmo_layers} show that the word representation layer of ELMo has can capture morphology faithfully while encoding little semantics. The semantic information is instead represented in the contextual layers. Comparing the results for embeddings extracted from different layers allows us to compare less semantically rich embeddings (non-contextualized word representations) with more semantically rich embeddings (contextualized embeddings). A comparison of these layers' output was made to discover the influence of this difference in semantic information.

\subsection{Data \& Experiment}
The comparison was performed using a Swedish pre-trained ELMo model \cite{elmo_pretrained}. The nouns and their gender labels were extracted from UD treebanks. The noun embeddings were generated using their treebank sentence as context. 
The output is collected at the word representation layer, the first contextualized layer, and the second contextualized (output) layer. This resulted in a set of a little over 10k embeddings for every layer, from which 10\% was randomly sampled and split as test data. The embeddings have a size of 1024, and the hidden layer size of the classifier has a size double the input size, 2048. The results for this comparison are shown in table 2.
\begin{table}[]
\renewcommand{\arraystretch}{1.3}
\caption{Accuracy and loss for different layers of the ELMO model }
\label{contextualized}
\centering
\begin{tabular}{@{}lll@{}}
\toprule
                          & Loss  & Accuracy \\ \midrule
Word representation layer & 0.168 & 93.82    \\
First layer   & 0.260 & 92.13    \\
Second layer  & 0.274 & 91.24    \\ \bottomrule
\end{tabular}
\end{table}
\subsection{Results}
An apparent decrease in accuracy and an increase in loss is observed when classifying the gender of the contextualized word embeddings. The added semantic and contextual information is not only unhelpful but even detrimental to the classifier's performance. Based on these results, it can be argued that the classifier uses very little semantic information for classifying grammatical gender and that the semantic information added in this experiment acted like noise. 

\section{Word embeddings from stripped corpus}
In the previous experiment, we have observed that the classifier does not seem to strongly rely on semantic data to classify grammatical gender in word embeddings. This observation would lead to the hypothesis that it relies on information on form and agreement instead. 

Form and agreement could be encoded in word embeddings through the noun's relation with agreed words. A neuter noun in Swedish would have a high co-occurrence with the neuter article 'ett', thus leading to a strong relationship between the vectors for the noun and the word 'ett'.

To test this hypothesis, embeddings have been created from a corpus that has all forms of agreement removed through the removal of articles and the stemming of all words.

fastText \cite{fasttext} was used to create Swedish word embeddings from a corpus consisting of all Swedish articles on Wikipedia. Another set of embeddings was created from the same corpus, but all articles were removed from the corpus. A third set of embeddings was created from the same corpus after stemming it with the Snowball stemmer \cite{snowball}. 
A classifier was trained on these embeddings in the same configuration as the previous experiments. The results can be found in table 3. 

The classifier only manages accuracy of $85.61\%$ on the embeddings from the no articles corpus. This is almost a $6\%$ drop caused by the missing information, which is very significant, considering the $70\%$ majority baseline. This indicates that the relationship between nouns and articles in word embeddings is a large part of what encoded information on grammatical gender in word embeddings for Swedish.

When classifying the stemmed embeddings, the accuracy falls to $84.66\%$. It could be argued this is in part due to the decrease in quality of the embeddings overall that comes with stemming the corpus. However, it is still a clear indicator that the formal information is a source of information for the classifier.

\begin{table}[]
\renewcommand{\arraystretch}{1.3}
\caption{Accuracy and loss for embeddings with different source corpora }
\label{table_example}
\centering
\begin{tabular}{@{}lll@{}}
\toprule
                          & Loss  & Accuracy \\ \midrule
Wikipedia corpus & 0.247 &  91.37  \\
Wikipedia corpus, no articles  & 0.368 & 85.61    \\
Wikipedia corpus, stemmed  & 0.397 &  84.66  \\ \bottomrule
\end{tabular}
\end{table}


\section{Conclusion}

Our exploration of grammatical gender in word embeddings has shown some overlap between how grammatical gender is encoded in different languages and that the degree of the overlap could be connected to language relatedness. It has also been shown through a comparison of embeddings from different layers of a contextualized embeddings model that adding semantic information to embeddings is detrimental to a grammatical gender classifier's performance. Lastly, by creating embeddings from a corpus that is stripped of information on form and agreement, it has been shown that a noun's form and relationship to gender specific articles are an essential source of information for grammatical gender in word embeddings. It has also shown that the model with stemming and no articles still performs better than chance.

\vspace{2mm} 


\Urlmuskip=0mu plus 1mu\relax
\bibliographystyle{acl_natbib}
\bibliography{extend_abstract}

\end{document}